\documentclass[]{spie}  

\usepackage{amsmath,amsfonts,amssymb}
\usepackage{graphicx}
\usepackage[colorlinks=true, allcolors=blue]{hyperref}
\usepackage{float} 

\title{Exploring Transfer Learning for Deep Learning \\Polyp Detection in Colonoscopy Images Using YOLOv8}

\author[a]{Fabian Vazquez}
\author[a]{Jose Angel Nuñez}
\author[b]{Xiaoyan Fu}
\author[a]{Pengfei Gu}
\author[a]{Bin Fu}
\affil[a]{Department of Computer Science, University of Texas Rio Grande Valley, Edinburg, TX, USA}
\affil[b]{The Second Affiliated Hospital of Fujian University of Traditional Chinese Medicine, Fuzhou, Fujian, China}

\begin{document} 
\maketitle

\begin{abstract}
Deep learning methods have demonstrated strong performance in objection tasks; however, their ability to learn domain-specific applications with limited training data remains a significant challenge. Transfer learning techniques address this issue by leveraging knowledge from pre-training on related datasets, enabling faster and more efficient learning for new tasks. Finding the right dataset for pre-training can play a critical role in determining the success of transfer learning and overall model performance. In this paper, we investigate the impact of pre-training a YOLOv8n model on seven distinct datasets, evaluating their effectiveness when transferred to the task of polyp detection. 
We compare whether large, general-purpose datasets with diverse objects outperform niche datasets with characteristics similar to polyps. In addition, we assess the influence of the size of the dataset on the efficacy of transfer learning. Experiments on the polyp datasets show that models pre-trained on relevant datasets consistently outperform those trained from scratch, highlighting the benefit of pre-training on datasets with shared domain-specific features.

\end{abstract}


\section{INTRODUCTION}
\label{sec:intro}  

\begin{figure} [ht]
   \begin{center}
   \begin{tabular}{c} 
   \includegraphics[height=5cm]{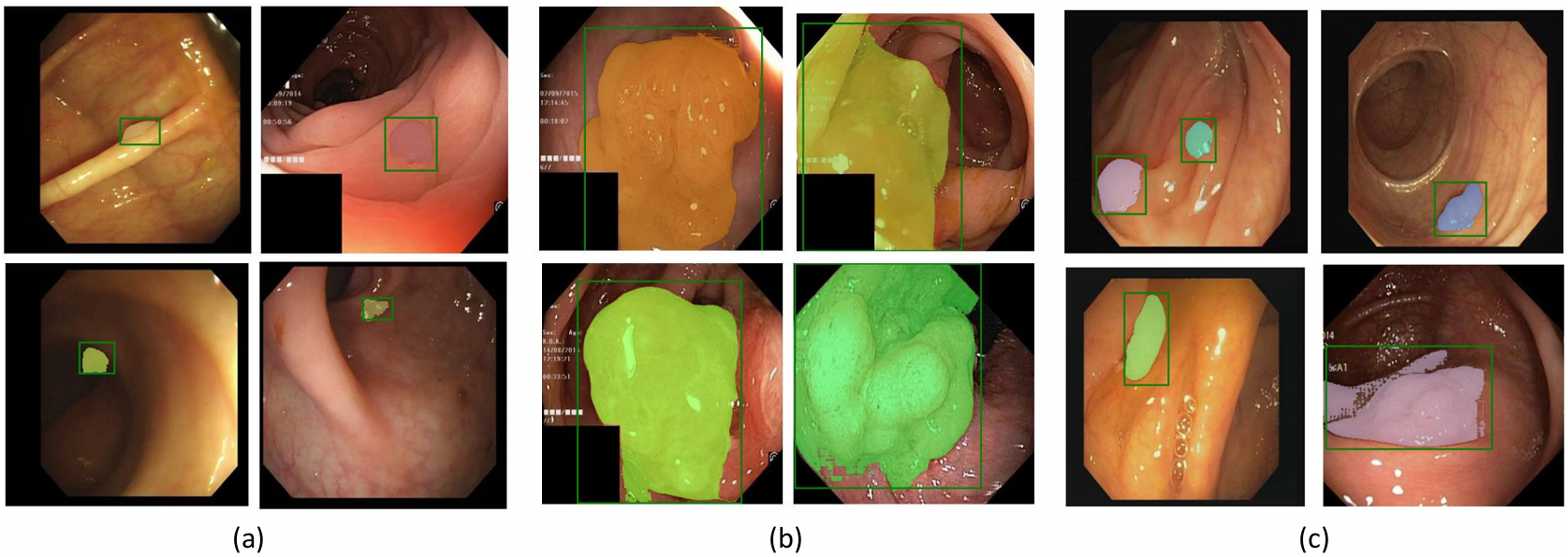}
   \end{tabular}
   \end{center}
   \caption[examples] 
   { \label{fig:examples} Visual examples highlighting challenges in detecting polyps: (a) small polyps, (b) extremely large or abnormal polyps, and (c) irregularly shaped or multiple polyps.}
   \end{figure}

Colorectal cancer is a significant health concern, being one of the leading causes of cancer-related mortality worldwide~\cite{arnold2017global}. Early detection of polyps during colonoscopy procedures is vital for preventing colorectal cancer. Studies have shown that early detection and removal of polyps can reduce the incidence and mortality of colorectal cancer by up to 90\%~\cite{cameron2023early}. Traditional methods for detecting polyps during colonoscopies rely heavily on the visual inspection skills of the endoscopist doing the exam. Despite the expertise of medical professionals, these methods are prone to variability and human error, leading to missed polyps, particularly small or flat ones. Variability in polyp size, shape, appearance, glare, and obstructions in colonoscopy images make the task much more challenging (see Fig.~\ref{fig:examples})~\cite{urban2018deep}.

Deep learning (DL) has made substantial advancements in recent years, particularly in computer vision tasks such as object detection~\cite{liang2020intracker}, classification~\cite{gu2024boosting}, and segmentation~\cite{gu2021k,gu2023convformer,zhang2023swipe,an2024sli2vol+}. One particularly DL model, YOLO (You Only Look Once), has grown in popularity due to its speed and efficiency~\cite{redmon2016you}. Various versions of YOLO have been made and are used for object detection, segmentation, and localization tasks. In~\cite{lalinia2024colorectal}, a YOLOv8 model was trained to detect polyps to observe its robustness and adaptability. Using polyp datasets such as the CVC-Clinic DB~\cite{bernal2015wm}, CVC-ColonDB~\cite{bernal2012towards}, ETIS-LaribPolypDB~\cite{silva2014toward}, and Kvasir-SEG~\cite{jha2020kvasir} for training and evaluation, Lalinia and Shafi were able to achieve high results with YOLO-v8 to show its effectiveness for real-time polyp detection in medical imaging. 
However, challenges remain in developing models that can generalize well across different domains and datasets. Issues such as dataset bias, variation in object appearance, occlusion, and complex backgrounds continue to affect the performance of computer vision systems~\cite{torralba2011unbiased}. Especially in object detecting tasks with limited data availability, such as in polyp detection. Additionally, real-time processing and the ability to operate effectively under varied and dynamic conditions are critical requirements for practical applications, including medical imaging~\cite{huang2017densely}.

Transfer learning is a technique that addresses the challenge of limited data availability by leveraging models pre-trained on large, diverse datasets to improve performance in specific tasks~\cite{pan2009survey}. This technique is especially valuable in medical applications, where acquiring large, annotated datasets is often costly and labor-intensive, as annotations typically require expert annotations from medical professionals. By using transfer learning, researchers can use pre-trained models to transfer useful features to new tasks, improving both model accuracy and training time~\cite{shin2016deep}. 
Datasets like ImageNet~\cite{deng2009imagenet} that contain over 14 million images and around 1000 classes, are commonly used to pre-train DL models in computer vision tasks~\cite{kornblith2019better}. 
In Ribeiro et al.~\cite{ribeiro2016exploring}, several convolutional neural network (CNN) models were pre-trained on ImageNet and then trained to do polyp classification. These models outperformed the models that received no pre-training and were only trained on the polyp classification task. In a similar study by Hon and Khan~\cite{hon2017towards}, transfer learning enabled DL models pre-trained on multiple different datasets to achieve state-of-the-art results related to medical imaging domains such as Alzheimer's disease classification from MRI images. In addition to detection and classification tasks, transfer learning has also proven effective in segmentation tasks such as segmenting vertebrae, the esophagus, lung lobes from CT scans, or prostate from MR scans~\cite{humpire2023transfer}. These researchers found in their study that using similar CT and MR scans for pre-training proved effective at improving the DL model's results when compared to no pre-training. Additionally, they tested their pre-trained model without fine-tuning (training on the testing task e.g. vertebrae from CT scans) and with fine-tuning and found that overall transfer learning with fine-tuning obtained higher results than their other two training strategies (pre-training only and scratch).

In this paper, we aim to examine the capabilities of transfer learning in improving model performance in polyp detection as a means to compensate for lack of data. To do so, we further investigate the impact of transfer learning on polyp detection performance to discover more optimal methods in training for polyp-related DL tasks that will reduce training time, data, and resources needed. Specifically: (1) Examine the effects of in-domain versus out-of-domain transfer learning by pre-training on datasets ranging from general objects (out-of-domain) to medical images (in-domain) with characteristics similar to polyps. (2) Evaluate whether larger pre-training datasets offer greater benefits compared to smaller datasets typically used for fine-tuning. Extensive experiments on four public datasets (CVC-Clinic DB~\cite{bernal2015wm}, CVC-ColonDB~\cite{bernal2012towards}, ETIS-LaribPolypDB~\cite{silva2014toward}, and Kvasir-SEG~\cite{jha2020kvasir}) reveal that models pre-trained on datasets with features analogous to polyps outperform those pre-trained on general objects, and consistently exceed the performance of models trained from scratch.

The contributions of this paper are the following: (1) Investigating the benefits on pre-training in-domain versus out-of-domain such as on COCO or medical datasets. (2) Releasing our pre-trained YOLOv8n models and code on our \href{https://github.com/fvazqu/yolo_training}{Github} so they can be used in relevant studies.

   \begin{figure} [ht]
   \begin{center}
   \begin{tabular}{c} 
   \includegraphics[height=8cm]{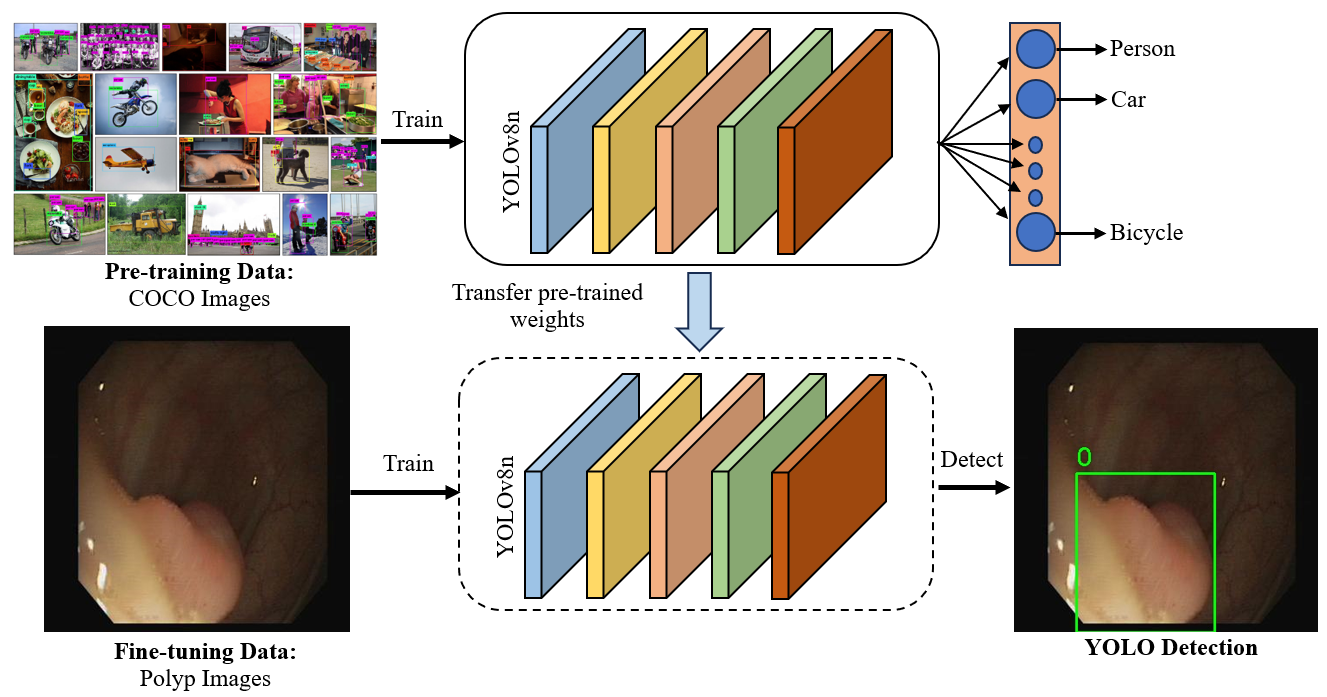}
   \end{tabular}
   \end{center}
   \caption[pipeline] 
   { \label{fig:pipeline} The overview of the transfer learning pipeline for deep learning-based polyp detection in colonoscopy images.}
   \end{figure}

\section{Methodology}
Fig. \ref{fig:pipeline} illustrates the transfer learning pipeline for deep learning-based polyp detection. The process includes pre-training various models on different datasets, fine-tuning them using a polyp detection dataset, and evaluating the models' performance in detecting polyps.

\subsection{Pre-Training} \label{pre-training}
In this stage, different YOLOv8n models were each pre-trained on different datasets, ranging from out-of-domain datasets such as the COCO dataset to in-domain datasets related to medical imaging. The aim of this pre-training phase was to equip the model with robust feature extraction capabilities that could later be fine-tuned for the domain-specific task such as polyp detection. Different YOLOv8n models are pre-trained on a respective pre-training dataset and are compared with a baseline model trained from scratch (without pre-training). The pre-training was carried out in two phases. 

In the first phase of pre-training, smaller datasets were used to train the YOLOv8n models. The pre-training datasets used are each divided into 80\% for training and 20\% for validation. The images were resized to $640 \times 640$ with pixel values normalized to a range of 0 to 1. No separate testing set was used in order to allocate more data for training. The datasets used for pre-training are as follows: (1) \textbf{The MRI Brain Tumor dataset:}~\cite{ahmed2023} This public dataset from Kaggle is used for bounding box detection of different brain tumors. The dataset contains 2,175 images and 4 classes and was split into 1,740 training images, and 435 validation images. The model trained on this dataset is named \textbf{YOLO-brain-tumor}. (2) \textbf{The Acne dataset:}~\cite{kurnaz2023} This public dataset from Kaggle is used for training to detect acne in images. It contains 927 images and only one class. It was split into 741 training images, and 186 validation images. The model trained on this dataset is named \textbf{YOLO-acne}. (3) \textbf{The Fruit and Vegetable Detection for YOLOv4 dataset:}~\cite{patel2023} This public dataset from Kaggle is used to detect 14 different types of fruits and vegetables from images. It contains 4,592 images and 14 classes. It is split into 3,673 training images, and 919 validation images. The model trained on this dataset is named \textbf{YOLO-fruit\&vegetable}.

Since the COCO17 dataset contains over 100,000 images~\cite{lin2014microsoft} which is significantly larger than most public datasets, we utilized other pre-training datasets that contained a minimum of 10,000 images or combined different datasets with the same type of images for the second phase of the pre-training. This was done to observe the effects of diversity and training size on transfer learning. The datasets used are each divided into 80\% for training and 20\% for validation. The images were resized to $640 \times 640$ with pixel values normalized to a range of 0 to 1. No separate testing set was used in order to allocate more data for training. The models trained in the second pre-training phase are denoted with an "-XL" to show they are trained on the larger datasets containing more than 10,000 images. The datasets used for the second phase of pre-training are as follows:

\textbf{The Skin Cancer: HAM10000 dataset} is a public dataset, accessed through Kaggle,and comprises of 10,015 images used for skin lesion segmentation~\cite{ghuwalewala2018}. This dataset was employed independently, without the Acne dataset used in Phase 1, due to significant differences in data types and classes. Given the limited availability of publicly accessible acne datasets specifically tailored for segmentation or object detection tasks, the Skin Cancer: HAM10000 dataset was selected based on its similar characteristics to acne. The model trained on this dataset is named \textbf{YOLO-HAM10000-XL}.

\textbf{The Combined MRI Brain Tumor dataset} is a dataset composed of five distinct datasets focused on brain tumor detections from MRI scan. The combined dataset resulted in 10,997 images. They are (1) The MRI Brain Tumor dataset~\cite{ahmed2023} : This dataset was used in phase one as well. (2) The MRI for Brain Tumor with Bounding Boxes dataset~\cite{sorour2023} : This public dataset is from Kaggle and consists of 2,625 images. (3) The Brain-Tumor dataset~\cite{nahin2021} : This public dataset is also found on Kaggle and consists of 1,101 images. (4) The Brain-Tumor(YOLO) dataset~\cite{pandey2023} : This public dataset is also found on Kaggle and consists of 3,064 images. (5) The Brian Tumor Dataset (Roboflow)~\cite{fate522024} : This public dataset is found on Roboflow and consists of 2,032 images. The model trained on this dataset is named \textbf{YOLO-brain-tumor-XL}.

\textbf{The Combined Fruit and Vegetables dataset} consists of two datasets that together make 13,071 images and 17 classes. (1) The Fruit and Vegetable Detection for YOLOv4 dataset~\cite{patel2023} : This dataset was used in Phase 1 for pre-training. (2) The Fruit Detection dataset~\cite{tyagi2023} : This public dataset from Kaggle consists of 8,479 images and is utilized for fruit detection. The model trained on this dataset is named \textbf{YOLO-fruit\&vegetable-XL}.

The following two models were used as controls to compare our newly pre-trained models. \textbf{The COCO17 dataset} is a public dataset that contains 123,287 images and 80 different classes of diverse objects~\cite{lin2014microsoft}. Ultralytics provides a default YOLOv8n model already trained on this dataset. As a result, no pre-training on the COCO17 dataset had to be done since the pre-trained model by Ultralytics was utilized for this paper. This model was trained on the COCO 2017 dataset by Ultralytics for 100 epochs and is denoted \textbf{YOLO-coco} in this study. Additionally, another model was used as the control to compare without pre-training. The \textbf{Trained from scratch} model had no pre-training, and was only fine-tuned as seen in the next section.

The in-domain datasets used in this paper are the HAM-10000 and the combined brain tumor datasets because of their similarities with the polyp images and being part of the medical field. The out-of-domain datasets are the COCO17 and the combined fruit and vegetable datasets because they are not precisely related to medical imaging. The COCO17 dataset was picked because of its use in previous studies related to transfer learning and to observe the effects of transfer learning with datasets consisting of a large amount of classes. The combined Fruit and Vegetables dataset was chosen for the similarities certain fruits have with polyps, such as shape and shine.

\subsection{Fine-Tuning} \label{fine-tuning}
The second stage involved fine-tuning the pre-trained YOLOv8n model on a polyp detection dataset. This process aimed to adapt the model's generalized feature representations to the specific task of detecting polyps in colonoscopy images. Fine-tuning was conducted using four publicly available polyp detection datasets: the CVC-ClinicDB~\cite{bernal2015wm} which contains 612 images, the CVC-ColonDB~\cite{bernal2012towards} which contains 380 images, the ETIS-LaribPolypDB~\cite{silva2014toward} which contains 196 images, and the Kvasir-SEG~\cite{jha2020kvasir} which contains 1000 images. These four datasets were combined and randomly split to use 80\% for training, 10\% for validation, and 10\% for testing. When combined, the polyp datasets yielded 1,748 training images, 220 validation images, and 220 testing images.`

The fine-tuning process was performed at intervals of 20, 50 and 100 epochs. This was done to measure the convergence speed during training and see if transfer learning benefitted training time. The hyperparameters were kept consist as with pre-training maintaining the same batch size, learning rate, and optimization settings. As in the pre-training phase, the same data augmentation techniques were applied to enhance robustness and prevent overfitting. Transfer learning was employed by initializing the YOLOv8n model with the weights obtained during pre-training and continuing the fine-tuning training.

The datasets for fine-tuning first required obtaining the bounding box annotations with the correct YOLO format from the ground truth segmentation masks of the polyp datasets~\cite{bernal2015wm}~\cite{bernal2012towards}~\cite{silva2014toward}~\cite{jha2020kvasir} in order to train the YOLOv8n models. The minimum and maximum contours of the ground truth masks were taken to calculate the center coordinates of the bounding box, box width, and box height. These were then saved in text files utilized for training YOLO. Post pre-training, the models were fine-tuned on the combined polyp dataset using a new YAML file. The same hyperparameters from pre-training were applied to the fine-tuning, as well as the data augmentation settings. The ‘pretrained’ argument was set to 'true' during the fine-tuning stage to enable the transfer of the learned weights. It was left 'false' during the pre-training stage.

\subsection{Evaluation} \label{eval}
After fine-tuning, the trained model was evaluated on the separate testing set created to assess the model's ability to detect polyps in colonoscopy images. The test set was derived from the same polyp detection datasets used for fine-tuning but was kept independent during the training process to ensure unbiased evaluation.

Performance was measured using several evaluation metrics, including precision, recall, F1-score, mean Average Precision (mAP) at a treshold of 0.5, and mAP at a treshold of 0.5 to 0.95. Precision assesses the ratio of the correctly identified polyps among all the instances where a polyp is detected, calculated as the sum of all correct detections and incorrect detections. This metric illustrates the percentage of correct predictions made by the model, and a higher precision value indicates fewer false positives. Recall, represents the ratio of correctly detected objects over all the objects it is supposed to detect, calculated as the sum of all the correct detections and missed detections. A high recall value is crucial in preventing missed detections. The F1-score is a combined metric that considers both recall and precision. It provides a more comprehensive evaluation of the model’s detection performance by balancing the trade-off between false positives and false negatives. As a result, a higher F1-score a model is more accurate. Lastly, the mean Average Precision (mAP) is a comprehensive metric used to evaluate the performance of object detection models. It is a common evaluation metric used in computer vision tasks where localization of the object is done with bounding boxes. This metric is calculated by finding the Average Precision (AP) for each class and the average over a number of classes. This is done by calculating the area under the precision-recall curve at the respective thresholds. The model was deployed to detect polyps in real-time, generating bounding boxes around detected polyps for each image in the testing set. These metrics are commonly used when evaluating computer vision models that involve detecting objects with bounding boxes, such as YOLO.

\section{Experiments and Results} \label{exp}
\label{sec:exp}

\subsection{Implementation Details}
In this paper, the YOLOv8n~\cite{yolov8_ultralytics} (nano) model was utilized, which comprises of 225 layers and approximately 3.01 million learning parameters. It is designed for efficient object detection. The libraries used in this paper are: Ultralytics YOLOv8.2.42 packag, Python-3.12.0, Pytorch version: 2.5.0.dev20240620+cu124, Torchvision version: 0.20.0.dev20240625+cu124, and CUDA 12.5. Training was done on a local computer with an NVIDIA GeForce RTX 4090 Laptop GPU and with Google Colab’s T4 GPU and A100 GPU. 

The pre-training process was carried out over 100 epochs with the following fixed hyperparameters set at default by Ultralytics: a batch size of 16, an initial learning rate of 0.002, and an adam optimzer\cite{kingma2014adam} as the optimization algorithm. Data augmentation techniques such as random cropping, horizontal flipping, and rotation were applied to improve model generalization. Upon completion of pre-training, the learned weights were saved for transfer to the fine-tuning phase. All experiments were conducted with the same training parameters across both stages, including consistent hyperparameters and identical data augmentation techniques.

\subsection{Experimental Results}
From all the results, the YOLO-fruit\&veg-XL model achieved the highest overall F1-Among all the models evaluated, the YOLO-fruit\&veg-XL model achieved the highest overall F1-score of 95.1\% after training for 100 epochs, as shown in Table \ref{tab:results_100b}. In experiments with YOLO models pre-trained on smaller datasets, the YOLO-coco model, pre-trained on the COCO17 dataset, consistently achieved the highest F1-scores at 20, 50, and 100 epochs. Its best performance was recorded at 100 epochs, reaching an F1-score of 94.5\%, as seen in Table \ref{tab:results_100s}. This result is likely due to the COCO dataset’s extensive diversity, complex backgrounds, numerous classes, and high-quality annotations.

The results further demonstrate that increasing the number of images used for pre-training enhances model performance when employing transfer learning. Specifically, when YOLO models were pre-trained on at least 10,000 images, their performance improved significantly, as evident in Tables \ref{tab:results_20b}, \ref{tab:results_50b}, and \ref{tab:results_100b}, where the XL models consistently outperformed their smaller counterparts. Among these models, YOLO-fruit\&veg-XL achieved the highest F1-scores at 50 and 100 epochs, while at 20 epochs, it was surpassed only by the YOLO-brain-tumor-XL model.

For models pre-trained on smaller datasets, the YOLO-fruit\&veg model ranked second at 20, 50, and 100 epochs, with the YOLO-coco model achieving the highest scores at each stage. Although in-domain datasets did not yield the best results, larger and more diverse datasets proved more effective for transfer learning. Notably, despite being considered out-of-domain for polyp detection, the YOLO-fruit\&veg dataset may have contributed to improved performance due to visual similarities between fruit, vegetables, and polyps in colonoscopy images.

\begin{figure}[H]
    \centering
    \includegraphics[width=1.0\linewidth]{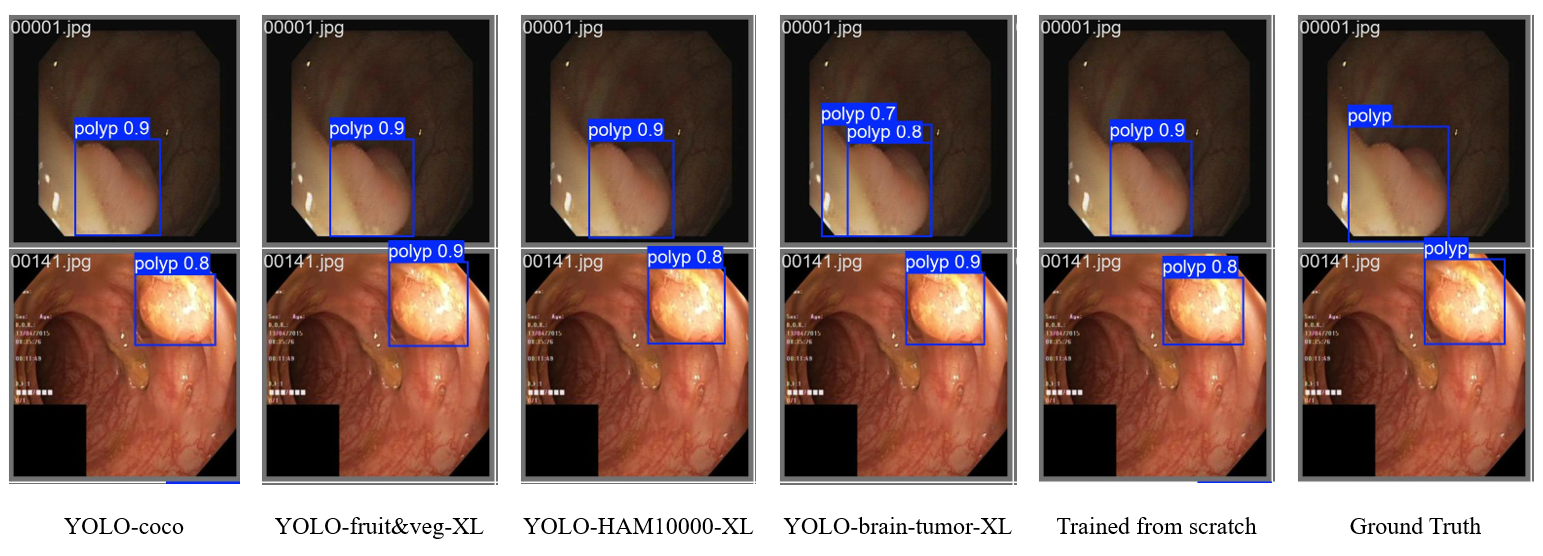}
    \caption{Visual results of polyp detections by different methods.}
    \label{fig:vis}
\end{figure}

Finally, the base YOLOv8n model, trained from scratch, consistently produced the lowest F1-scores. This underscores the effectiveness of pre-training in improving model training and performance. Additionally, pre-trained models learned faster and required less training time to achieve high F1-scores, as observed in Table 10. The analysis of model performance on polyp detection demonstrates a slight improvement when models are pre-trained on in-domain datasets containing relevant and transferable features when compared to no pre-training. Furthermore, incorporating greater diversity and complexity in pre-training datasets appears to enhance overall model performance. Notably, datasets with the highest number of classes, such as the COCO17 and the combined Fruit and Vegetables dataset, achieved the highest scores overall, showing the value of diversity in pre-training datasets. Although the combined Fruit and Vegetables dataset is significantly smaller than the COCO17 dataset, it was able to improve the performance slightly, perhaps due to common similarities with the fine-tuned task. In Fig. \ref{fig:vis}, the models were evaluated on the polyp testing dataset with detections visualized using bounding boxes to indicate the presence of polyps. The figure provides a comparison between multiple model predictions and the ground truth for polyp localization.

\begin{table}[H]
\caption{Polyp detection results for models pre-trained on smaller datasets at 20 epochs.} 
\label{tab:results_20s}
\begin{center}
 \scalebox{0.7}{
\begin{tabular}{|l|c|c|c|c|c|} 
\hline
\rule[-1ex]{0pt}{3.5ex}  Model            & Precision (\%) & Recall (\%) & F1-score (\%) & mAP50 (\%) & mAP50-95 (\%)\\
\hline
\rule[-1ex]{0pt}{3.5ex}  YOLO-fruit\&veg  & 90.5     & 83.3  & 86.8    & 91.7 & 66.1\\
\hline
\rule[-1ex]{0pt}{3.5ex}  YOLO-acne        & 89.9     & 80.3  & 84.8    & 90.2 & 61.5\\
\hline
\rule[-1ex]{0pt}{3.5ex}  YOLO-brain-tumor & 87.6     & 83.8  & 85.7    & 91.1 & 63.7\\
\hline
\rule[-1ex]{0pt}{3.5ex}  YOLO-coco        & 90.6     & 88.3  & 89.4    & 94.0 & 68.6\\
\hline
\rule[-1ex]{0pt}{3.5ex}  Trained from scratch & 83.5 & 81.8  & 82.6    & 87.4 & 59.8\\
\hline
\end{tabular}
}
\end{center}
\end{table} 

\begin{table}[H]
\caption{Polyp detection results for models pre-trained on smaller datasets at 50 epochs.} 
\label{tab:results_50s}
\begin{center}
 \scalebox{0.7}{
\begin{tabular}{|l|c|c|c|c|c|} 
\hline
\rule[-1ex]{0pt}{3.5ex}  Model            & Precision (\%) & Recall (\%) & F1-score (\%) & mAP50 (\%) & mAP50-95 (\%)\\
\hline
\rule[-1ex]{0pt}{3.5ex}  YOLO-fruit\&veg  & 90.9     & 92.0  & 91.4    & 95.9 & 72.0\\
\hline
\rule[-1ex]{0pt}{3.5ex}  YOLO-acne        & 87.8     & 85.1  & 86.4    & 92.5 & 68.1\\
\hline
\rule[-1ex]{0pt}{3.5ex}  YOLO-brain-tumor & 89.3     & 84.1  & 86.6    & 92.3 & 65.2\\
\hline
\rule[-1ex]{0pt}{3.5ex}  YOLO-coco        & 90.2     & 89.9  & 90.0    & 95.5 & 70.2\\
\hline
\rule[-1ex]{0pt}{3.5ex}  Trained from scratch & 90.5 & 87.5  & 88.9    & 93.1 & 67.1\\
\hline
\end{tabular}
}
\end{center}
\end{table} 

\begin{table}[H]
\caption{Polyp detection results for models pre-trained on smaller datasets at 100 epochs.} 
\label{tab:results_100s}
\begin{center}
 \scalebox{0.7}{
\begin{tabular}{|l|c|c|c|c|c|} 
\hline
\rule[-1ex]{0pt}{3.5ex}  Model            & Precision (\%) & Recall (\%) & F1-score (\%) & mAP50 (\%) & mAP50-95 (\%)\\
\hline
\rule[-1ex]{0pt}{3.5ex}  YOLO-fruit\&veg  & 93.4     & 93.4  & 93.4    & 95.3 & 73.6\\
\hline
\rule[-1ex]{0pt}{3.5ex}  YOLO-acne        & 95.4     & 93.0  & 94.2    & 96.3 & 74.1\\
\hline
\rule[-1ex]{0pt}{3.5ex}  YOLO-brain-tumor & 93.4     & 93.4  & 93.4    & 96.5 & 73.8\\
\hline
\rule[-1ex]{0pt}{3.5ex}  YOLO-coco        & 95.3     & 94.3  & 94.5    & 96.1 & 73.1\\
\hline
\rule[-1ex]{0pt}{3.5ex}  Trained from scratch & 93.9 & 92.1  & 92.9    & 95.7 & 72.4\\
\hline
\end{tabular}
}
\end{center}
\end{table} 

\begin{table}[H]
\caption{Polyp detection results for models pre-trained on larger datasets at 20 epochs.} 
\label{tab:results_20b}
\begin{center}     
 \scalebox{0.7}{
\begin{tabular}{|l|c|c|c|c|c|} 
\hline
\rule[-1ex]{0pt}{3.5ex}  Model            & Precision (\%) & Recall (\%) & F1-score (\%) & mAP50 (\%) & mAP50-95 (\%)\\
\hline
\rule[-1ex]{0pt}{3.5ex}  YOLO-fruit\&veg-XL  & 92.9     & 89.0  & 90.9    & 94.5 & 70.7\\
\hline
\rule[-1ex]{0pt}{3.5ex}  YOLO-HAM10000-XL    & 92.9     & 84.2  & 88.3    & 92.9 & 68.9\\
\hline
\rule[-1ex]{0pt}{3.5ex}  YOLO-brain-tumor-XL & 92.3     & 90.8  & 91.5    & 95.1 & 70.2\\
\hline
\rule[-1ex]{0pt}{3.5ex}  YOLO-coco           & 90.6     & 88.3  & 89.4    & 94.0 & 68.6\\
\hline
\rule[-1ex]{0pt}{3.5ex}  Trained from scratch & 83.5 & 81.8  & 82.6    & 87.4 & 59.8\\
\hline
\end{tabular}
}
\end{center}
\end{table} 

\begin{table}[H]
\caption{Polyp detection results for models pre-trained on larger datasets at 50 epochs.} 
\label{tab:results_50b}
\begin{center}     
 \scalebox{0.7}{
\begin{tabular}{|l|c|c|c|c|c|} 
\hline
\rule[-1ex]{0pt}{3.5ex}  Model            & Precision (\%) & Recall (\%) & F1-score (\%) & mAP50 (\%) & mAP50-95 (\%)\\
\hline
\rule[-1ex]{0pt}{3.5ex}  YOLO-fruit\&veg-XL  & 95.9     & 92.4  & 94.1    & 96.7 & 74.7\\
\hline
\rule[-1ex]{0pt}{3.5ex}  YOLO-HAM10000-XL    & 94.7     & 88.2  & 91.3    & 95.4 & 72.7\\
\hline
\rule[-1ex]{0pt}{3.5ex}  YOLO-brain-tumor-XL & 93.5     & 90.4  & 91.9    & 95.9 & 71.7\\
\hline
\rule[-1ex]{0pt}{3.5ex}  YOLO-coco        & 90.2     & 89.9  & 90.0    & 95.5 & 70.2\\
\hline
\rule[-1ex]{0pt}{3.5ex}  Trained from scratch & 90.5 & 87.5  & 88.9    & 93.1 & 67.1\\
\hline
\end{tabular}
}
\end{center}
\end{table} 

\begin{table}[H]
\caption{Polyp detection results for models pre-trained on larger datasets at 100 epochs.} 
\label{tab:results_100b}
\begin{center}  
 \scalebox{0.7}{
\begin{tabular}{|l|c|c|c|c|c|} 
\hline
\rule[-1ex]{0pt}{3.5ex}  Model            & Precision (\%) & Recall (\%) & F1-score (\%) & mAP50 (\%) & mAP50-95 (\%)\\
\hline
\rule[-1ex]{0pt}{3.5ex}  YOLO-fruit\&veg-XL      & 95.9     & 94.3  & 95.1    & 97.1 & 74.2\\
\hline
\rule[-1ex]{0pt}{3.5ex}  YOLO-HAM10000-XL        & 95.8     & 92.1  & 94.4    & 97.0 & 74.9\\
\hline
\rule[-1ex]{0pt}{3.5ex}  YOLO-brain-tumor-XL     & 94.6     & 91.9  & 93.2    & 96.0 & 72.6\\
\hline
\rule[-1ex]{0pt}{3.5ex}  YOLO-coco               & 95.3     & 94.3  & 94.5    & 96.1 & 73.1\\
\hline
\rule[-1ex]{0pt}{3.5ex}  Trained from scratch    & 93.9 & 92.1  & 92.9    & 95.7 & 72.4\\
\hline
\end{tabular}
}
\end{center}
\end{table}

\subsection{Challenges and Limitations}
Our methods has several limitations. One key limitation is data availability, particularly the lack of sufficient datasets with YOLO-format annotations. Additionally, the experiment was conducted exclusively with the YOLOv8n "nano" network, despite the existence of other versions, such as "s," "m," "l," and "x" \cite{yolov8_ultralytics}, as well as alternative CNN and transformer-based networks that could provide valuable comparisons. Computational constraints were another challenge, including the training time required and the difficulties in hyperparameter tuning. Furthermore, the number of images used for pre-training was limited, as obtaining a large and diverse dataset with consistent annotations for specific classes is inherently challenging. Lastly, the metrics used in this paper, such as the F1-score, may not fully capture the clinical relevance of the model or how effectively it can assist medical practitioners in real-world scenarios.

\subsection{Future Work}
Further adjustments to the training could have been experimenting with the hyperparameters such as the batch size, data augmentations, learning rate, number of epochs, and optimizer. Trying different data augmentation techniques could improve data quantity and generalization, which could potentially lead to better results because certain augmentations may be more beneficial for certain datasets. Additionally, testing the model’s performance over multiple epochs is crucial for optimization, such as trying out a different number of epochs to observe the effects of training on the different models with different pre-training datasets. Furthermore, exploring alternative models beyond YOLOv8, including transformer-based architectures or other convolutional neural networks (CNNs), could provide valuable insights to see what. These avenues of exploration hold promise for advancing the reliability, accuracy, and generalizability of DL applications in medical imaging. A very important method for accurate testing includes using the k-fold cross validation, which is widely used for the evaluation of many machine learning tasks. Future research could benefit from employing k-means cross-validation to ensure more reliable and statistically robust testing outcomes, minimizing biases that may arise from single-split evaluations. Expanding pre-training efforts by incorporating additional datasets with varied complexities and characteristics can also enhance the model’s ability to transfer knowledge effectively for tasks such as polyp detection to avoid issues such as overfitting. Which may have contributed to the reason the model’s performance achieved high results on the polyp datasets used for testing, since the fine-tuning dataset is relatively small and consists of multiple images of the same polyp from different perspectives. Another aspect that could be explored is exploring different pre-training methodologies, such as pre-training with the COCO17 dataset first and then continuing the pre-training with a smaller dataset. This method could allow datasets to compensate for the disparity in size when comparing with very large datasets such as the COCO17 dataset. Lastly, due to the lack of publicly available annotated polyp datasets, future efforts to increase the data in this domain will allow for models to be evaluated on larger datasets and improve robustness and polyp detection.

\section{Conclusions}
This paper demonstrates the impact of transfer learning on polyp detection by using YOLOv8 and pre-training different models on separate datasets that incorporate a diverse amount of classes and objects or task-specific features with two classes. The pre-trained YOLO models consistently outperformed the model trained from scratch at all epochs, showing how pre-training is effective at improving training and model performance. The "XL" pre-trained models trained on more data consistently outperformed their counterparts that were trained on smaller datasets. This shows that datasets with greater diversity and more data offer more advantages for transfer learning and overall performance improvement. Although the in-domain datasets did not achieve the highest results, the model trained on the combined Fruit and Vegetable dataset achieved the highest F1-scores and mAP values compared to the model trained on the COCO17 dataset. This shows that shared similarities can also improve performance by transferring knowledge (weights) across tasks. This is particularly crucial for tasks with limited public data available, such as in polyp detection, where pre-training datasets can compensate for the lack of data. Despite the improved performance, there remains a gap in utilizing more datasets to accurately assess the models as well as employing more reliable testing results such as k-folds cross validation in order to validate any conclusions made. Perhaps developing a system to quantify the effectiveness of transferring the weights from pre-training for a specific task could be explored to make transfer learning and training more efficient. This research contributes to the potential of transfer learning in enhancing medical image analysis, paving the way for more accurate and reliable diagnostic tools. Other contributions include sharing the pre-trained YOLO-models, the processed datasets utilized in this paper such as the Ham-10000 dataset with YOLO bounding box annotations, the combined MRI brain tumor dataset, and the combined fruit and vegetables dataset containing more than 10,000 images. The code will also be made available to help advance future studies.

\section{Acknowledgements}
This work was supported in part by the Graduate Assistance in Areas of National Need (GAANN) grant.

\bibliography{report} 
\bibliographystyle{spiebib} 

\end{document}